# Description Logics based Formalization of *Wh*-Queries

Sourish Dasgupta
DA-IICT, India
sourish_dasgupta
@daiict.ac.in

Rupali KaPatel
DA-IICT, India
rupali.it.08
@gmail.com

Ankur Padia
DA-IICT, India
padiaankur
@gmail.com

Kushal Shah
DA-IICT, India
kushalshah40
@daiict.ac.in

## ABSTRACT
The problem of Natural Language Query Formalization (*NLQF*) is to translate a given user query in natural language (NL) into a formal language ($\mathcal{F}$) so that the $\mathcal{F}$ semantic interpretation has equivalence with the NL interpretation. Formalization of NL queries enables logic based reasoning during information retrieval, database query, question-answering, etc. Formalization also helps in Web query normalization and indexing, query intent analysis, etc. In this paper we are proposing a *Description Logics* based formal methodology for *wh*-query intent (also called *desire*) identification and corresponding formal translation. We evaluated the scalability of our proposed formalism using Microsoft Encarta 98 query dataset and OWL-S TC v.4.0 dataset.

## Categories and Subject Descriptors
I.2.4 [**Knowledge Representation Formalism and Methods**]: Representation (*procedural and rule-based*)

## 1. INTRODUCTION
*Natural Language Query Formalization* (*NLQF*) is a formal and systematic procedure of translating a user query in natural language into an expression in a formal language without losing the semantics of the user query. The choice of a formal language can range from SQL used in conventional DBMS to more advanced SPARQL for RDF graph based databases designed to support the Semantic Web. However, any translation process involves rigorous linguistic analysis of NL queries and constructing a formal semantic interpretation that is equivalent to the original query semantics. NLQF has a twofold effect. First, it helps in proper identification of the query intent (also called *desire*). Secondly, it formally defines the query intent in relation to other linguistic constituents of the query thereby providing a platform for logical reasoning based semantic information retrieval and question answering [1-2]. The biggest challenge involved in NLQF is to efficiently and accurately identify the innate desire and to understand the linguistic nuances during a translation process. A query may bear the same semantic content but may be structurally different. For an example, the query: "*Who is the greatest crime novel writer?*" is semantically equivalent to the query: "*What is the name of the greatest author of crime novel?*". It may also have the same grammatical structure but bear different semantics.

NLQF has been a topic of intensive research in the database community. Most of the effort was concentrated in translating NL queries into formal representations suitable for database retrieval such as SQL [3-4]. More recent researches have shown a growing interest in mining graph databases represented in RDF-like format. As a result several works have been proposed to translate NL query into SPARQL-like formalism [5-7, 18]. However, many of these works support at the most shallow lexico-syntactic query analysis extracting heuristic patterns which are then translated into SPARQL like queries. Many other approaches are ontology based where an external set of ontologies are required for mapping query tokens to the most probable formal concept (mostly RDF/RDFS/OWL represented) so as to link together the mapped tokens into a formal semantic graph structure (such as SPARQL, nRQL) [5-6, 18]. The graph structure is then matched with similar graphical representations of document content for query answering. However, ontology assisted NLQF heavily depends on the correctness and completeness of the external ontologies and may not be very accurate if the target information source is independent of the imported ontology set. Moreover, such RDF databases are nothing more than very light-weight knowledge bases with no high-end reasoning support required for knowledge discovery. Hence, if the target corpus is a formal knowledge base then SPARQL cannot serve as a suitable formal query language.

Other *distributional hypothesis* [8] based purely statistical approaches has also been proposed mostly for NL query processing which can hardly fall under NLQF [9-11]. One of the intrinsic problems of statistical approaches is that query goal (or desire/intention) detection is very difficult if linguistic analysis is ignored. Also information retrieval largely depends on similarity measure models that are mostly token co-occurrence based [12-13]. Such co-occurrence analysis cannot guarantee semantic similarity with the query goal.

In this paper we propose a deep linguistic analysis based semantic formalization framework for NL *wh*-queries in English. DL representation of queries provides the support to perform formal subsumption based reasoning over DL based knowledgebase for knowledge discovery. We show that such queries can be neatly characterized into a syntactic structure, called *Query Characterization Template* (*QCT*), covering most possible linguistically valid query variations. The primary aim of QCT is to identify the query desire and the relationship of the desire with the query input. This leads to the next step of accurate query formalization. However, such characterization is non-trivial and involves capturing positional nuances of query tokens correctly. We also show that a Description Logic (DL) [25] sub-language exists that has semantic equivalency with that of *wh*-queries. We have presented the salient rules for NL query to DL query translation. The proposed methodology is independent of any external ontology assistance. The scope of this paper is limited to *wh*- queries of six kinds: (i) *what*, (ii) *which*, (iii) *who*, (iv) *when*, (v) *where*, and (vi) non-procedural *how*. Our contribution in this paper is as follows:

**Table 1**[1]

| NLQP System | NLQ Type | FL | Ont. Aided | Lex. Aided | Target Corps | L.A. |
|---|---|---|---|---|---|---|
| [10] | UR | No | No | No | ATIS | TAN |
| **LASSO** [9] | UR | Keyword based patterns | No | Word Net | NL docs | L-SPL |
| **AquaLog** [7] | R C.Voc | Triple based | Yes | Word Net | Sem. mark-up docs | No |
| **Power-Aqua** [24] | UR | Triple based | Yes many | Word Net | Distr. sem. docs | No |
| **Querix** [6] | UR | Triple based SPARQL | Yes | Word Net | Selected set of Ont. | L-SPL |
| **PANTO** [5] | UR | Triple based SPARQL | Yes | Word Net | Ont. | POS TAG |
| [23] | UR | DL based $\mathcal{ALC}$ | Yes | NA | Ont. | POS TAG |
| **Masque/ SQL** [3] | R | SQL | No | No | RDBMS | PTr. |
| **PRECISE** [4] | UR | SQL | No | from DB | ATIS | PTr. |
| **START** [21] | UR | NL Ann. based | No | MIT Lexicon | Web corpus | L-SPL |
| QCT based[2] | UR | DL based | No | Word Net | NL docs | POS TAG |

1. A novel query desire-input dependency analysis theory, termed QCT, is proposed.
2. Proposing DL has a suitable candidate formal semantic theory for query formalization.
3. Evaluation in terms of characterization accuracy using Microsoft Encarta query dataset and query dataset built on OWL-S TC v.4.0 dataset.

The paper is organized into the following sections: (i) related work outlining some of the major contributions in NL query processing, (ii) problem statement defining the problem of NLQF formally, (iii) Approach where query characterization and DL formalization has been discussed at length, and (iv) Evaluation in terms of characterization accuracy.

## 2. RELATED WORKS

Various approaches for NL Query Formalization can be broadly classified into two main categories: (i) statistical learning based analysis, and (iii) lexico-syntactic analysis. Table 1 is an overview of various approaches for developing NLQP systems. We have categorized them on the basis of various parameters which differentiate them. One of those parameters is *query nature* which can be of two types: (i) *restricted* (*R NL*), and (ii) *unrestricted* (*UR NL*). Restricted NL based systems cannot accept queries of all linguistics forms and hence, provide query formulation only for NL queries that can be given through some sort of controlled vocabulary. *Ontology aided* systems import external ontologies as input for aiding NL queries into their respective formal representations. *Lexicon aided* systems use lexicons (or thesauri) for enriching NL query vocabulary which in turn aids in normalized query formalization. By *target corpus* we mean the resource from which the answer is expected. We see that systems in this respect can be either NL document corpus based or ontology based.

In [10] author has tried to detect goal (i.e. *desire*) from the user's NL query using *Tree-Augmented Naive Bayes* networks (*TANs*) for goal detection but this work is domain specific and semantic values are compromised here. In [14] a conversion tool which takes queries expressed in NL and an ontology as input and returns the appropriate formal queries. It uses WordNet to disambiguate the words and triple based model for formalization. The generated queries are then sent to the reasoner for querying the knowledge bases. In this work ontology to be queried is required to be chosen by user and it doesn't support complex query formulation. In [15] an approach to Semantic Information Retrieval of semantically annotated documents, based on NL understanding of query, has been proposed. This work incorporates an OWL query ontology for SPARQL based inference. In [16] a nested *CG* (*Conceptual Graph*) language for formal representations of natural language queries has been proposed. In [17] a formal semantic analysis of object queries required for the modern object-oriented databases has been proposed. Unlike other object query languages, a number of realistic features including object identity, object creation and invocation of methods that need not terminate has been covered. A translation procedure from NL query into a formal language query such as SPARQL has been described in [18]. In this paper a user query is translated into SPARQL by choosing the most appropriate query from the prepared queries. Queries for the knowledge base and a set of corresponding normalized queries for the problem has been prepared beforehand and user's query is mapped to one of query which is obtainable from the knowledge base. For relatively large knowledge base such WWW this may not be scalable approach. In [19] models of DP services as RDF views over a mediated (domain) ontology has been proposed. Each RDF view contains concepts and relations from the mediated ontology to capture the semantic relationships between input and output parameters. Query rewriting algorithms for processing queries over DP services and query mediator which automatically transforms a user's query (during the query rewriting stage) into a composition of DP services.

*START* [20] was the first online question-answering system which uses statistical NLP techniques and lexico-syntactic pattern matching. Another system called *NLP-Reduce* [21] was proposed which is also based on lexico-syntactic pattern matching. In this system query keywords are mapped with synonym enhanced triple stores in the target corpus. A key feature was that user queries can be phrases or complete queries and also need not be grammatically correct. A domain-independent system, called *Querix*, was proposed in [6] where full English questions had to be given which were then parsed for extracting triple patterns. These triples are extracted out of a prior query skeleton which is generated based on word categories. These triple patterns are then mapped onto the target knowledge base for match. A guided input NL search engine, called *Ginseng*, was proposed in [22].

---
[1] NLQP: Natural Language Query Processing; FL: Formal Language; L-SPL: Lexico-Syntactic Pattern Learning; C. Voc.: Controlled Vocabulary; PTr.: Parse Tree; Ann.: Annotation; DB: Database; Ont.: Ontology; Lex: Lexicon; Corps: Corpus; L.A.: Linguistic Analysis; Sem.: Semantic

[2] proposed DL based framework

## 3. PROBLEM OVERVIEW

The problem of NLQF involves 3 core tasks:

**Task 1**: To choose a formal grammar $\mathcal{G}$ that can generate the dependency structure (ex: parse tree) of the linguistic components of a given NL query accurately.

**Task 2**: To select a formal language $\mathcal{F}$ that has an interpretation $\mathcal{F}_\mathcal{J}$ such that there exists a one to one correspondence with the semantic interpretation of the formal grammar $\mathcal{G}$.

**Task 3**: To model a representational equivalence function $\tau_{\mathcal{L}_{NL}}$ that takes in a query Q in $\mathcal{L}_{NL}$ and maps it to an expression $\mathcal{E}$ in $\mathcal{F}$ such that maximum semantic preservation is achieved. Such semantic preservation can be achieved through model-theoretic semantic constructions. $\tau_{\mathcal{L}_{NL}}$ should be consistent and should not generate expressions that are mutually inconsistent.

Task 1 involves syntactic parsing of all valid NL queries. This is a challenging task since NL queries can be of varied forms with (a) similar syntactic structures but different semantics (ex: "*Where is the capital of Florida?*" - Answer: "*28.5N, 81.3W*" vs. "*What is the capital of Florida?*" - Answer: "*Orlando*") and (b) similar semantics but different syntactic structures (ex: "*Where is the capital of Florida?*" vs. "*What is Orlando's location?*" - Answer for both: "*28.5N, 81.3W*"). In the former case the queries should be characterized such that they have their corresponding formal language translation unique while in the latter case both the queries should have the same normalized formal language translation. Moreover, queries may not always be simple (i.e. no clausal constraint; ex: "*Who are Alexandar's favorite Greek mythological heroes?*") but can be complex (ex: "*Who are the heroes of Greek mythology who were Alexander's favorite?*") and compound ("*Who are the heroes of Greek mythology and Alexander?*"). Although structurally the queries are simple, complex and compound yet semantically they are the equivalent. Hence, they should have the same answer: "*Achilles*". This requires accurate part-of-speech tagging (POS tagger) and parse tree generation based on which a given NL query is fitted into a chosen characterization grammar.

Task 2 consists of translation of the parsed queries into a formal language representation (usually the model-theoretic style). Just like task 1 formal translation is also not trivial since it involves: (i) resolution of several ambiguities and linguistic nuances including re-formulation and normalization of semantically equivalent NL queries having different structures, (ii) computational query resolution (ex: "*How far is New York from Orlando?*"), (iii) comparative/superlative query resolution (ex: "*What is the highest mountain in Asia?*"). The scope of this paper is limited to the *wh*-queries of the kinds: (i) *what/which*, (ii) *who/whom/whose*, (iii) *where*, (iv) *when*, and (v) *how much*.

### 3.1 Problem Definition

Given an NL *wh*-query $Q_{Wh}^{NL}$ in English model a transformation function $\tau_{Q_{Wh}}$ such that:

- $\tau_{Q_{Wh}}: Q_{Wh}^{NL} \mapsto Q_{Wh}^{FL}$
- $\left(Q_{Wh}^{NL}\right)^{I_{NL}} \equiv \left(Q_{Wh}^{NL}\right)^{I_{FL}}$

where:

- $Q_{Wh}^{FL}$ is the formalization of $Q_{Wh}^{NL}$ in the formal language $\mathcal{F}$
- $I_{NL}$ is the linguistic reading of English
- $I_{FL}$ is the semantic interpretation function of $\mathcal{F}$

## 4. APPROACH

### 4.1 Characterization (Task 1)

Any user query has two primary linguistic components - (i) *desire* (or intent) of the query and (ii) *input* of the query. While the query desire is essentially what an answer needs to satisfy, the query input provides the satisfiability constraint on the desire. For an example, in the query: "*What is the capital of USA?*" while the desire is an instance of the entity Capital, the input is USA acts as a constraint imposed on the instance relating it to USA (and not just any other country/province). For task 1 (described in section 3.1) choosing a formal grammar theory serves as a basis for generating the parse tree of a query. Constructing semantics by applying a particular semantic theory over complicated (although sophisticated) parse trees is computationally expensive. We observed that since our chosen semantic theory is Description Logics (DL), where concepts are defined in terms of roles and their associations with other concepts (thus, forming *subject-predicate-object* triples), the primary objective of query parsing should be to identify the desire (which is the subject of the query), predicate, and input (which is the object of the query). Hence, a full-fledged parse tree is not necessary for the case. We developed a pseudo-grammar structure, called *Query Characterization Template* (*QCT*), that captures the intrinsic *desire-input dependency structure* in all forms of English factual queries (simple, complex, and compound). This dependency generates a unique QCT for each of the three forms. QCT is a pseudo-grammar in several senses: (i) the sequence of the lexicons of the original query can get changed after the characterization process, (ii) lexicons can get normalized into a standard form after characterization, and (iii) it does not give a generic set of rules to combine or split phrase structures but rather "*fits in*" queries of equivalent grammatical structure into one fixed template. We hereby define the *wh*-query and its three forms of as follows:

**Definition 1** (*Wh-Query*): A *Wh*-Query is a query that contains at least one of the following query tokens (or their equivalent lexical variations): *what, which, who, whose, whom, when, where, how, why*.

*What* query can be: (i) definitional such as "*What is a cat?*" (expected answer: class definition), (ii) inclusion such as "*What animals are mammals?*" (expected answer: sub classes of *mammal*); (iii) instance retrieval such as "*What is the capital of USA?*" (expected answer: a *city* instance), (iv) class retrieval such as "*What is Taj Mahal?*" (expected answer: class of instance *Taj Mahal*), and (v) instance associated concept retrieval such as "*What does John drink in the morning?*" (expected answer: sub classes of *drink* that *John* has for morning).

*Which* query is similar to *What* queries except that such queries cannot be definitional.

*Who* query behaves sometimes as a *Which* query and sometimes as a *What* query with the special underpinning that the expected answer is related to either a named animal, or a person (or person group/organization).

*When* query can be: (i) absolute temporal such as "*When is Thanksgiving?*" (expected answer: a particular day of a month), and (ii) relative temporal such as "*When will John arrive?*" (expected answer can be: after/before some event).

*Where* query can be: (i) absolute spatial such as "*Where is the leaning tower of Pisa?*" (expected answer: geographical location), and (ii) relative spatial such as "*Where is the ball?*" (expected answer can be: on/under/below/at some object).

*How* query can be: (i) procedural such as "*How is a flan made?*"(expected answer: recipe or step-wise set of actions), (ii) state based such as "*How is Joe?*" (expected answer: current health status of Joe), (iii) quantitative such as "*How much does the bag cost?*" (expected answer: price), or (iv) computational such as "*How far is Tampa from Miami?*" (expected answer: computed distance). We do not consider procedural *how* in the scope of this paper.

*Why* query is causal in nature such as "*Why is the grass green?*". We also leave out *why* queries from the scope of this paper.

**Definition 2** (Simple *Wh*-Query): A Simple *Wh*-Query is a *wh*-query that consists of a single and non-clausal query desire (explicit or implicit) and a single, unconstrained, and explicit query input.

Example simple *wh*-query is "*What is the capital of USA?*". In this case the desire (*Capital*) is explicit, single, and unconstrained by any clausal phrase. The input (*USA*) is also explicit, single, and unconstrained by any clausal phrase. It should be noted that, as remarked earlier, the input is an implied constraint over the desire which is different than clausal constraint. Also, the desire may be implicit sometimes. For an example, in the query: "*What is a tomb?*" the implicit desire is the definition of *Tomb* (i.e. description of the class *Tomb*) while the unconstrained and single input is *Tomb*.

**Definition 3** (*Complex Wh-Query*): A Complex *Wh*-Query is a *wh*-query that consists of a single query desire (explicit or implicit) and multiple explicit query input.

**Definition 4** (*Complex Non-Clausal Wh-Query*): A Complex Non-Clausal *Wh*-Query is a complex *wh*-query that is clausal constraint free on both the query desire and the multiple query input.

Example complex non-clausal query is "*In which country is the state capital of Missouri located?*". In this query the desire *Country* is unconstrained. There are two input (*State Capital* and *Missouri*) each of which is also unconstrained.

**Definition 5** (*Complex Clausal Wh-Query*): A Complex Clausal *Wh*-Query is a complex *wh*-query that consists of at least one clausal constraint on either the query desire or query input or both.

Example complex clausal *wh*-query is "*Who was the British Prime Minister who was elected two times one of which was during World War II?*". In this query the single explicit desire is the (instance of the) class *British Prime Minister* having no clausal constraint. There are two query input: *two times* and *World War II*. Also, the first input *two times* has a clausal constraint "*one of which was during …*".

**Definition 6** (*Compound Query Wh-Query*): A Compound *Wh*-Query is a *wh*-query that consists of conjunctive/disjunctive lexicons between one or more simple *wh*-queries or complex *wh*-queries.

Example of compound *wh*-query is "*What are the available car models of Volkswagen and their respective prices?*". In the following sub-sections the QCT of each of the three forms of sentences has been discussed at length.

### 4.1.1 QCT of Simple Wh-Query

A simple *wh*-query can be characterized according to the following structure:

$$[Wh][[R_1]][[[Q]^*][[M]^*][D]][[R_2]][[[Q]^*][[M]^*][I]][?]$$

where:

$[[\ ]]$: second square bracket indicates optional component

$[D]$: Query desire class/instance - value restricted to {NN, NNP, JJ, RB, VBG}[3]

$[I]$: Query input class/instance - value restricted to {NN, NNP, JJ, RB, VBG}

$[R_1]$: Auxiliary relation - includes variations of the set {*is, is kind of, much, might be, does*}

$[R_2]$: Relation that acts as (i) predicate of $D$ as the subject and $I$ as the object or (ii) action role of $I$ as the actor - value restricted to {VB, PP, VB-PP}[1]

$[[Q]^*]$: Quantifier of $D$ or $I$ - values restricted to {DT}[1]. The * indicates that $Q$ can recur before $D$ or $I$.

$[[M]^*]$: Modifier of $D$ or $I$ - value restricted to set {NN, JJ, RB, VBG}. The * indicates that $M$ can recur before $D$ or $I$.

We can observe that this QCT can cover all the linguistically valid 180 questions (excluding quantifiers and modifiers) according to the given definition of simple wh-query. $R_1$ is auxiliary role in the sense that it cannot act as a predicate of either the $D$ or the $I$. However, $R_1$ serves as a good indicator for resolving several linguistic ambiguities. For an example, in a *how* query if $R_1$ is *much* (or its lexical variations) then it is a quantitative query while in a *who* query if $R_1$ is *does* (or its lexical variations) then the associated verb is an activity (i.e. Gerund; ex: "*Who does singing?*" - *Singing* is an activity in this case).

$R_2$ is a relation that can either be associated with $D$ as the subject or $I$ as the subject but not both. If $R_2$ is positioned after $D$ in the original query then $R_2$'s subject is $D$. For an example, in the simple query "*What is the capital of USA?*" the subject of $R_2$ (*of*) is $D$ (*Capital*) and the object is $I$ (*USA*). If $R_2$ is positioned after $I$ in the original query then its subject is $I$. For an example, in the query "*Which country is California located in?*" the subject of $R_2$ (*located in*) is $I$ (*California*) and object is $D$ (*Country*). Table 2 lists some of the important simple *wh*-query characterization.

#### 4.1.1.1 Implicit Desire Identification

Implicit query desire implies that $D$ is empty. This can happen if and only if the following query structures are found:

1. $[Wh][is/does][[\ ]][\ ][[Q]][[M]^*][I][?]$

---

[3] Abbreviations follow the conventions of Penn Treebank POS tags. [30]

**Table 2**

| Natural Language Wh-Simple Query | Wh-Simple Query Characterization |
|---|---|
| What is the capital of Gujarat? | [Wh] = 'What', [$R_1$] = 'is', [D] = 'the capital', [$R_2$] = 'of', [I] = 'Gujarat', [?] |
| Which is the highest mountain in world? | [Wh] = 'Which', [$R_1$] = 'is', [D] = 'the highest mountain',[$R_2$] = 'in', [I] = 'world', [?] |
| How many legs does a millipede have? | [Wh]='How many', [D] = **count**('legs'), [$R_2$]= 'does have', [I] = 'millipede', [?] |
| What are some dangerous plants? | [Wh] = 'What', [$R_1$] = 'are',[I] = 'dangerous plants', [?] |
| Where is California? | [Wh] = 'Where', [$R_1$] = 'is', [I] = 'California', [?] |

**Table 3**

| | |
|---|---|
| What is most populous democracy in the Caribbean which is geographically the largest as well? | [Wh]=what,[$R_1$]=is,[D]=the_most_populous_democracy, [$R_k$]=in,[$I_1^k$]=the_Caribbean, [$Cl_{k+1}$]=which, [$R_{k+1}$]=is, [$I_1^{k+1}$]=geographically_the_largest? |
| What is the distance between Missouri and Texas? | [Wh]=what,[$R_1$]=is [D]=the_distance ,[$Cl_k$]=null,$R_k$=between, [$I_1^k$]=Misssouri, [CC]=and, [$I_2^k$] =Texas? |

2.  [Wh][ ][[ ]][[$R_2$]][[Q]][[M]*][I][?]
3.  [Wh][$R_1$][[ ]][[$R_2$]][[Q]][[M]*][I][?]

If $R_1$ does not exist then *D* is empty. If $R_2$ does not exist while $R_1$ exists then *D* is empty. For an example, in the query: "*What is converted into diamond?*" $R_1$ is identified to be *is* by default and $R_2$ is detected to be *is converted into*. However, there is no lexicon in between $R_1$ and $R_2$ (structure 3). Therefore, *D* is empty. Another case in which *D* always remains empty is when the *wh*-query is a *where* or a *when* query. This also holds true for complex and compound queries. For an example, in the query: "*When is the next solar eclipse?*" the query characterization is as:

[When][is][[D = TIME]][R = null][the][next][solar][eclipse][?]

### 4.1.1.2 Explicit Desire Identification

As an extension to the observation the previous section we can conclude that any lexicon between $R_1$ and $R_2$ is *D*. For an example, in the simple query "*What is the capital of USA?*" $R_1$ is identified to be *is* and $R_2$ is detected to be *of*. Therefore, D is *Capital*. After *D* is identified the remaining lexicon is *I*.

### 4.1.2 QCT of Complex Wh-Query

A complex *Wh*-query can be characterized according to[4]:

[Wh] [[$R_1$]] [[D]] [[$Cl_D$]][[$R_2$]] [ $I_1^1$ ] [([CC][ $I_2^1$ ])*] …

… [[$Cl_2$]][[$R_3$]] [ $I_1^2$ ] [([CC][ $I_2^2$ ])*] …

… [[$Cl_N$]] [[$R_{N+1}$]] [ $I_1^N$ ] [([CC][ $I_2^N$ ])*] [?]

---

[4] Modifiers and quantifiers are not associated with *D* and *I*. They are associated in exactly the same way as QCT of simple queries.

**Table 4**

| Natural language Compound Wh-Query | Compound Wh-Query Characterization |
|---|---|
| What happens when you mix potassium permanganate and glycerin? | [$Wh^i$]=what,[$R_1^i$]=null,[$D_1^i$]=happens(implicit activity),[$Cl_1^i$]=when, [$R_2^i$]=mix, [$I_1^i$]=potassium permanganate,[CC]=and,[$I_2^i$]= glycerine, [CC] ...=null? |
| How long will an electric car run and how fast can it go? | [$Wh^i$]=Howlong,[$R_1^i$]=will,[$D_1^i$]=count(implicit), [$Cl_1^i$]=null, [$R_2^i$]=mix, [$I_1^i$]=potassium permanganate,[CC]=and,[$I_2^i$]= glycerine? |
| What is shape and size of baloon when air comes out? | [$Wh^i$]=What,[$R_1^i$]=is,[$D_1^i$]=shape, [CC]=and,[$D_2^i$]=size, [$Cl_1^i$]=null,[$R_2^i$]=of,[$I_1^i$]=baloon, [$Cl_2^i$]=when, [$R_3^i$]=comes_out,[$I_2^i$]=air,[CC] ...= null? |
| What is the travelling charge to Bombay and hotel_rent in Bombay? | [$Wh^i$]=What,[$R_1^i$]=is,[$D_1^i$]=the_travelling_charge, [CC]=and,[$D_2^i$]=hotel_rent,[$Cl_1^i$]=null,[$R_2^i$]=in,[$I_1^i$]=bombay ,[CC] ...=null? |
| Who were the foremost authorities in discovering algebraic formulas, theorems, and/or expressions? | [$Wh^i$]=Who,[$R_1^i$]=were,[$D_1^i$]=the_foremost_authorities, [$Cl_1^i$]=null,[$R_2^i$]=in_discoverying, [$I_1^i$]=,algebraic_formulas, [CC] = and, [$I_2^i$] =theorems [CC] = and,, [$I_3^i$]= expressions,[CC] ...=null? |
| Which volcanoes are active and which is which ones are dormant? | [$Wh^i$]=Which[$R_1^i$]=null,[$D_1^i$]=volcanoes, [$Cl_1^i$]=null,[$R_2^i$]=are_active,[$I_1^i$]=,null,[CC] = and, [$Wh^j$]=Which[$R_1^j$]=null,[$D_1^j$]=volcanoes, [$Cl_1^j$]=null,[$R_2^j$]=dormant,[$I_1^j$]=,null? |

where:

$Cl_D$: clausal lexicon (constraining *D*)

$Cl_2$: second clausal lexicon (constraining $I_l$)

$Cl_k$: clausal lexicon associated with structure [ $I_1^k$ ] [([CC][ $I_2^k$ ])*]

[CC] : conjunctive/disjunctive lexicon for *I*

[D] : query desire - value restricted to {NN, NNP, JJ, RB, VBG}

[$I_l^k$] : *l*-th query input *k*-th structure - value restricted to {NN, NNP, JJ, RB, VBG}

$[R_{k+1}]$: relation associated with the k-th clause that acts as (i) predicate of *D* as the subject and *I* as the object or (ii) action role of *I* as the actor - value restricted to {VB, PP, VB-PP}

$[[M]^*]$: modifier of the *D* or the *I* - value restricted to set {NN, JJ, RB, VBG}. The * indicates that M can recur before *D* or *I*.

In this QCT we see the possible repetition of the structure $\left[I_1^k\right]\left[([CC][I_2^k])^*\right]$. Within this structure there is an optional sub-structure $\left[([CC][I_2^k])^*\right]$ that may add to the number of input within each of such structures. A clausal lexicon in a complex clausal *wh*-query is always associated with such a structure. The number of clausal lexicons is the same as the number of such structures in a given query. It should be noted that there must be at least two such structures for a query to qualify as complex. Also, clausal lexicons in the general case is optional and hence, the QCT also holds true for complex non-clausal *wh*-query. We name the following structure as *clausal structure* (*CS*):

$[[Cl_D]][[R_2]] [I_1^1] [([CC][I_2^1])^*]$ ...

... $[[Cl_2]][[R_3]] [I_1^2] [([CC][I_2^2])^*]$ ...

... $\left[[[Cl_N]] [[R_{N+1}]] [I_1^N] [([CC][I_2^N])^*]\right] [?]$

Example complex query characterization is given in table 3. The given QCT can cover 1800 linguistically valid complex queries (excluding quantifiers and modifiers).

### 4.1.3 QCT of Compound Wh-Query

A compound *Wh*-query can be characterized according to:

$[Wh^1]\left[[R_1^1]\right]\left[[D_1^1][([CC_D][D_2^1])^*]\right]\left[[CS^1]\right]$ ...

... $\left[CC_Q^1\right][Wh^2]\left[[R_1^2]\right]\left[[D_1^2][([CC_D][D_2^2])^*]\right]\left[[CS^2]\right]$ ...

... $\left[CC_Q^{N-1}\right][Wh^N]\left[[R_1^N]\right]\left[[D_1^N][([CC_D][D_2^N])^*]\right]\left[[CS^N]\right][?]$

where:

$[CC_D]$ : Conjunctive/disjunctive lexicon for *D*

$[CC_Q]$ : Conjunctive/disjunctive lexicon for *wh*-sub-query

Compound query characterization example has been given in table 4.

## 4.2 DL as Formal Query Language (Task 2)

In our approach we choose the formal language $\mathcal{F}$ to be Description Logics (DL). As mentioned earlier we argue that most factual IS-A sentences have expressive equivalency in the DL language: $\mathcal{AL}[\mathcal{U}][\mathcal{E}][\mathcal{H}][\mathcal{J}][\mathcal{O}](\mathcal{D})$ where:

$\mathcal{AL}$: Attributive Language – supports atomic concept definition, concept intersection, full value restriction, limited role restriction, and atomic concept negation.

$[\mathcal{U}]$: Union – supports concept union

$[\mathcal{E}]$: Existential – supports full role restriction

$[\mathcal{C}]$: Complement – supports concept negation

$[\mathcal{H}]$: Role Hierarchy – supports inclusion axioms of roles

$[\mathcal{O}]$: Nominal – supports concept creation of unrecognized Named Entity

$[\mathcal{J}]$: Role Inversion - supports inverse roles

$(\mathcal{D})$: Data Type – supports range concepts to be data type

The choice of DL over other semantic theories has several reasons: (i) DL is equivalent to the guarded $\mathcal{L}_2$ fragment of FOPL and hence, is decidable [25], (ii) DL representation is compact and variable-free as compared to representations such as DRS [26] and LFT [27] making it comparatively easy to parse, (iii) the DL sub-language $\mathcal{AL}[\mathcal{U}][\mathcal{E}][\mathcal{H}][\mathcal{J}][\mathcal{O}](\mathcal{D})$ is tractable since we observed that most IS-A sentence interpretation is covered by $\mathcal{AL}[\mathcal{U}][\mathcal{E}][\mathcal{J}][\mathcal{O}]$, (iv) highly optimized semantic tableau based DL reasoners [28] are available as compared to slower hyper-resolution based theorem provers used in DRS or LFT based reasoning, (v) DL has direct mapping with the W3C recommended OWL format for web ontology[5]. Expressions in DL can represent two types of queries: (i) general queries such as "*What is a synagogue?*" (answer is a T-Box definition or inclusion axiom in the knowledgebase), and (ii) specific queries such as "*What is the name of the highest mountain in Australia?*" (the answer is an A-Box assertion in the knowledgebase).

## 4.3 DL Formalization (Task 3)

As mentioned in the previous section, NL queries can be of two types in the context of DL: (i) *T-Box queries* and (ii) *A-Box queries*. T-Box queries can be: (i) definitional (ex: "*What is a cat?*", (ii) inclusion (ex: "*What animals are mammals*?"), and (iii) super class retrieval (ex: "*What kind of animal is lion?*". A-Box queries on the other hand can be: (i) instance retrieval ("*Who resides in 221B Baker Street?*", (ii) class retrieval (ex: "*Who is Agatha Christie?*", and (iii) instance associated concept retrieval ("*What does John drink in the morning?*"). Some queries are ambiguous and the linguistic reading may imply either T-Box definitional or A-Box instance retrieval (ex: "*Who is a student?*" - Answer 1: "*John and Joe are students*"; Answer 2: "*A student is a person who studies in an educational institution.*") We argue that correct and complete DL formalization of query implies that query processing (and hence, question-answering) can be formulated as either a T-Box subsumption reasoning or an A-Box retrieval reasoning over a knowledgebase. We do not include $[\mathcal{C}]$ (i.e. concept negation) in this work since we exclude from the scope of this paper formalization of queries with negative clauses (such as "*What is an animal called that cannot lay egg?*").

### 4.3.1 Base Translation Rules

As discussed in section 4.1 we model any *wh*-query to have two components - desire and input. We also mentioned that QCT helps to establish *desire-input* dependency. From a DL formalization point of view such dependency identification naturally culminates to the DL *definition* of the desire in terms of the input. By definition we mean the model theoretic semantic interpretation of the description of a desire as constrained by the input. Given any simple *wh*-query Q having *D*, *I*, $R_2$ the following translation rules always holds true:

**Base Rule 1.1**: If $R_2$ is empty and *I* is not NNP or quantified then $D_F^S \sqsubseteq I$; $D_F^W \sqsupseteq I$;

**Base Rule 1.2**: If $R_2$ is empty and *I* is NNP then

$D_F^S \sqsubseteq WordNet.getMSP(I)$; $D_F^W \sqsupseteq WordNet.getMSP(I)$;

otherwise:

$D_F^S \sqsubseteq \{I\}$; $D_F^W \sqsupseteq \{I\}$;

**Base Rule 2.1**: If subject of $R_2$ is *D* and $R_2$ is not empty then $D_F = D \sqcap \exists R_2.I$;

---

[5] OWL DL is equivalent to $\mathcal{SHOIN}(\mathcal{D})$ while OWL 2 is equivalent $\mathcal{SROIQ}(\mathcal{D})$.

**Base Rule 2.2**: If subject of $R_2$ is $I$ and $R_2$ is not empty then $D_F = D \sqcap \exists R_2^{-1}.I$;

**Base Rule 3.1**: If subject of $R_2$ is $S$ and $I$ is NNP then

$D_F = D \sqcap \exists R_2 . WordNet.getMSP(I)$; otherwise:

$D_F = D \sqcap \exists R_2 . \{I\}$;

**Base Rule 3.2**: If subject of $R_2$ is $I$ and $I$ is NNP then

$D_F = D \sqcap \exists R_2^{-1}.WordNet.getMSP(I)$; otherwise:

$D_F = D \sqcap \exists R_2^{-1}.\{I\}$;

where:

$D_F$: Formalized desire

$D_F^S$: Strongly formalized desire

$D_F^W$: Weakly formalized desire

$D$: Desire component identified in QCT

$I$: Input component identified in QCT

$R_2$: Relation component identified in QCT that is associated with $D$ and $I$

*WordNet.getMSP*: A method developed to get the most specific parent class from WordNet v 2.1.

Base rule 1.1 is meant for T-Box queries in general except when the input is quantified (ex: "*Who is the student?*"). Strongly formalized desire ($D_F^S$) is an inclusion/definitional T-Box query and requires more specific answers (i.e. sub-classes of $I$). Weakly formalized desire ($D_F^W$) is an generic T-Box query and can allow less specific answers (i.e. super-classes of $I$ is allowed).

Base rules 2.1 and 2.2 are meant for A-Box queries. At an A-Box level the query formalism for rules 2.1 and 2.2 is: $D_F(?x)$; where *?x* is the variable that belongs to the class $D_F$. Rules 3.1 and 3.2 are meant for class retrieval queries and instance associated concept retrieval. Also rule 1.2 is class retrieval as well.

All the above base rules can be extended automatically for complex and compound queries as well. The core extension rules are discussed in the next section.

### 4.3.2 Extension Translation Rules

In this paper we discuss extension rules: (i) effect of modifiers, (ii) effect of clausal phrases, and (iii) effect of conjunctive and disjunctive phrases.

#### 4.3.2.1 Effect of Modifier

Normally, if a modifier in *wh*-query is a JJ or an NN then it modifies either an NN or an NNP. For an example, in the query: "*Who are the tall students?*" the JJ *Tall* modifies the input concept *Student* which is an NN. In such general cases it is evident that the concept *TallStudent* is a sub concept of the concept *Student*. An interesting phenomenon that can be observed for desire/input modification is what we term as *recursive nested modification*. In sentences where the subject modification is by a sequence of modifiers such as $[M_1][M_2][M_3][D/I]$ then a nested structure is assumed as: $[M_1]([M_2]([M_3]([D/I])))$ Here '( )' denotes scope of the modifier. Therefore, the scope of the inner most nested modifier $M_3$ is the concept $D$. The scope of the modifier $M_2$ is the sub-concept $M_3D$ formed as a result of the $M_3$ modifying $D$. At the same time $M_2$ also recursively modifies $D$ to form the sub-concept $M_2D$. Similarly $M_1$ has the sub-concept $M_2M_3D$ as scope of modification while in recursion modifies $M_3D$ and $D$. The T-Box rule for such recursive nested modification is as follows:

**Extension Rule** (Recursive Nested Modification: 3-level nesting):

$M_1D \sqsubseteq D;\ M_2M_1D \sqsubseteq M_1D;\ M_2M_1D \sqsubseteq M_2D;$

#### 4.3.2.2 Effect of Clausal Phrases

Complex *wh*-queries can be formalized by extending the base rule and extended rules of simple *wh*-queries. While formalization it is important to identify that whether the clausal constraint(s) is applied to desire or inputs. If it is an input constraint then which of the multiple inputs it is applied. This leads to a very important issue called *query dependency problem*. Query dependencies can be broadly classified as:

***Desire Dependency***: In some clausal complex *wh*-query constraint is applied on the desire. For example, in the query *"Which atomic bomb was dropped in Japan which had caused million people to die?"* the desire is name or type of atomic bomb with constraint: the bomb caused million people to die and was dropped in Japan. If clausal phrase contains an attribute of the desire then we assume it is constraint on desire. The given example query is characterized as:

$[Wh] = Which, [D] = atomic\ bomb, [R_1] = was, [R_2] =$
$dropped\ on, [I_1] = Japan, [Cl_1] = which, [R3] =$
$had\ caused\ to\ die, [I_2] = millions\ people, [?] = ?.$

Here type of atomic bomb is desire and clausal constraint (i.e. atomic bomb causing millions people to die). Therefore, constraint is considered to be applied on desire, not on input. Attributes are associated with relations of lexical variations of the structure {'*DESIRE which has*', '*DESIRE which includes*', '*DESIRE which is a*'}.

***Input Dependency***: If clausal phrase contains an attribute of the input then we assume it is constraint on input. For example, in the query *"What is the price of SLR camera which has 3.2 megapixel resolution?"* is characterized as:

"$[Wh] = What, [R_1] = is, [D] = the\ price, [R_2] = of, [I_1] =$
$SLR\ camera, [Cl_1] = which, [R_3] = has, [I_2] =$
$3.2\ megapixel\ resolution?$"

Here, "*3.2 megapixel resolution*" is a constraint input which is attribute of input "*SLR camera*". Attributes are associated with relations of lexical variations of the structure {'*INPUT which has*', '*INPUT which includes*', '*INPUT which is a*'}. In this section generic DL transformation rule for all complex *wh*-queries are given. All constraints can be formulated as intersection of concepts/instance given in the query.

**Extension Rule 1** (Complex Query: Inclusion T-Box):

$D_F^S \sqsubseteq \left( \sqcap_1^i I_i \right);$

An example query that requires this rule for translation is: "*What are the kinds of animals which are vegetarians?*". The corresponding equivalent DL is $D_F \sqsubseteq (Animal \sqcap Vegeterian)$.

**Extension Rule 2** (Complex Query: Input Dependency):

$D_F = D \sqcap \exists R_2. \left( I_1 \sqcap \exists R_2. \left( ... \sqcap (\exists R_{n+1}.I_n) \right) \right);$

**Extension Rule 3.1** (Complex Query: Desire Dependency):

$D_F = D \sqcap \exists R_2.I_1 \sqcap ... \sqcap \exists R_{n+1}.I_n;$

An interesting observation that we make is that if $R_1$ is empty while $D$ is non empty and constrained then $R_3$'s subject is $D$. For an example, in the query "*What country which is in Europe has the largest population?*" $R_3$ (*has*) has subject $D$ (*country*) since $R_1$ is empty. In this case the extension rule is as follows:

**Extension Rule 3.2** (Complex Query: Empty $R_1$):

$D_F = D \sqcap \exists R_2.I_1 \sqcap \exists R_3.I_1$;

### 4.3.2.3 Effect of Conjunctive/Disjunctive Phrases
To formalize the compound *wh*-queries, after the characterization process it is important to identify whether conjunctive/disjunctive phrase is applied on desire, input or relation. Compound queries can sometimes be split into simple queries and/or complex queries. They can then be formalized using simple query and complex query translation rules. We have defined the rules for cases when a given compound query $Q_{compound}$ can be split into conjunction of simple *wh*-queries. We have also done exhastive analysis of all possible structure of compound query structure by applying conjunctive/disjunctive lexicons between inputs, desires, relation. We concluded on 14 different forms of compound query.

The main motivation to break the compound query into simple query is to increase the precision and recall of the knowledge discovery system. If we can break the compound query into separated simple queries then later on all separated queries can be fired in parallel and answer of all separated queires can be returned by applying union operation between them. We have also defined the cases where splitting is not possible and separate formalization rules have been defined for them. More details on this topic is beyond the scope of this paper due to lack of space.

### 4.3.3 Non-Trivial Translation Rules
There are some queries whose semantic interpretation cannot be completely and correctly constructed in a straightforward way by applying a formal semantic theory. This is because of certain innate linguistic nuances that these queries carry that demand additional modification in the formal semantic representations. In the following sub sections we look into a few of such cases.

#### 4.3.3.1 Problem of Empty Input
In some queries the input can be empty. For an example, in the query "*Who barks?*" $R_2$ is a non-transitive verb with no explicit input as object. In such situation we need to do a reification of $R_2$ into its corresponding gerund sense and normalize the given NL query to the form [Wh][does][[ ]][ ][$f_r(R_2)$][?] where $f_r$ is the reification function. In the given example we can reformulate the query as "*Who does barking?*" and the corresponding DL rule is:

**Extension Rule** (Empty Input): $D := \exists does.Barking$

#### 4.3.3.2 Problem of Desire Inclusion
Some inclusion queries may have desire that have may have an intersection with input. For an example, in the query "*What kind of a water vehicle is also an air vehicle?*". In such cases the base rule 1.1 is modified as:

**Extension Rule** (Desire Inclusion): $D_F \sqsubseteq (D \sqcap I)$;

#### 4.3.3.3 Problem of Quantitative how-Query
In *how* queries that are quantitative in nature (i.e. $R_1 = \{much, many, \text{etc}\}$) we need to introduce a primitive concept **Count** and a primitive role **hasCount** where for any arbitrary satisfiable concept $C_i$ the following axiom holds: $C_i \sqsubseteq \exists hasCount.Count$. The **hasCount** is mapped to a function called $f_{count}$ that calculates the size of the instances of $C_i$ at any given point of time. For an example, in the query "*How many people live in New York?*" the count operator works on the desire *people* living in New York. The corresponding rule is:

**Extension Rule** (Quantitative *how*):

$D_F := Count \sqcap (\exists hasCount^{-1}.(D \sqcap \exists R.I))$;

#### 4.3.3.4 Problem of Temporal Adverbial Modifier
Some queries have temporal adverbial tokens such as in the query "*What can be sometimes observed in the morning sky?*" where $R_2$ (*observed in*) is associated with a temporal adverbial modifier (*sometimes*). The problem with formalizing such queries is that the ontological validity of the desire is essentially temporal in nature. In other words, for the given example, if a particular planet is observed in the morning sky it is not so that it will always be observed (like the sun which we can observe every day). Hence, sun cannot be a candidate answer in this case. The rule for such queries is as follows:

**Extension Rule** (Temporal Adverbial: *sometimes*):

$D_F = \exists sometimesR_2.I_1$;

$\exists sometimesR_2.I \sqcap \exists alwaysR_2.I = \bot$;

$\exists sometimesR_2.I \sqsubseteq \exists R_2.I; \exists alwaysR_2.I \sqsubseteq \exists R_2.I$;

#### 4.3.3.5 Problem of Superlative Modifier
In some queries superlative tokens are included such as in "*What is the tallest mountain in Europe?*". In such queries the desire is for a specific instance that has the optimal (maximal or minimal) degree of measurable modifier of the desire class. In the example *tall* is a measurable modifier whose superlative form is maximal height of all instances of the desire class *Mountain*. The *height* attribute of *mountain* is implicit in the given query. Keywords such as *most* and *least* are good indicators of deciding whether the computation has to be maximal or minimal. However, for suffix based superlative tokens (i.e. *est*) it is not so evident. The problem is how to know that *tall+est* has to maximized while *low+est* has to be minimized. We take a bootstrapping based approach with a seed bag of measurable modifiers (such as *tall*, *long*, *big*, *low*, *high*, *large*, *wide*, etc) and then mapped the bootstrapped keywords with corresponding plausible attributes (denoted $A_M$). For an example we get pairs such as: tall → {height}, big → {dimension, count}, wide → {breadth, width}, long → {length}, low → {depth}). Based on such pairing we then classify the modifiers into *positive modifiers* (those that requires maximization such as *tall*, *wide*, etc.) and *negative modifiers* (those that requires minimization such as *low*). The corresponding extension rule is:

**Extension Rule** (Superlative Queries): $D_F = f_{OPT}(C)(Integer \sqcap (\exists hasValue^{-1}.(A_M \sqcap \exists hasA_M^{-1}.(D \sqcap \exists R_2.I)))$;

where:

$f_{OPT}(C)$: Optimality function that returns *Integer* Datatype ($\mathcal{D}$).

## 5. EVALUATION
### 5.1 Evaluation Goal and Metric
Our evaluation aim was to observe the accuracy of the proposed QCT. We leave the accuracy evaluation of the DL formalization as a future work since that requires indirect comparative testing in terms of mean average precision and recall on some of the cutting-edge knowledge discovery systems. However, it is to be understood that the accuracy of the DL formalization is intrinsically dependent on the accuracy of QCT.

To evaluate QCT we decided on a simple *Characterization Coverage* (*CC*) measure. The measure is modeled to understand how many different linguistic forms of simple, complex and compound *wh*-queries in English can be identified correctly by

QCT. We measure CC in three perspectives: (i) *CC-Precision*, (ii) *CC-Recall*, and (iii) *CC-F1 score*. We define them as follows:

**CC-Precision**: Given a test set of NL queries the CC-Precision is calculated as the ratio of the number of correctly identified queries ($N_{CI}$) and the total number of identified queries in the test set ($N_I$).

**CC-Recall**: Given a test set of NL queries the CC-Recall is calculated as the ratio of the number of correctly identified queries ($N_{CI}$) and the total number of queries in the test set (N).

**CC-F1**: The Simple Harmonic Mean of CC-Precision and CC-Recall is the CC-F1.

## 5.2 Experimental Results

To evaluate CC of proposed work we have used the Microsoft Encarta 98 query test set [29] and OWL-S TC dataset. The Microsoft Question Answering Corpus (MSQA), which is aimed at querying documents belonging to the Encarta-98 encyclopedia. The test set contains 1365 usable English *wh*-queries. We excluded the queries of procedural *how* and *why* from this dataset. We have categorized simple, complex and compound queries from the dataset. There are total 473 queries of procedural *how* and *why* which are excluded. The reduced dataset consist of total 982 queries, which is distributed among 676 simple, 147 complex and 69 compound wh-queries. The accuracy statistics is given in table 5. We observe that the CC-Precision is 100% for all types of *wh*-queries while the overall CC-Recall is 94.50. The perfect precision shows that the QCT is theoretically sound.

To validate our results with Encarta 98 dataset we also tested QCT on custom query dataset built on top of OWL-S TC v.4.0 dataset[6]. The OWL-S TC dataset consists of service descriptions of 1083 web services from 9 different domains. A service description is a formal specification of the behavior of a web service in terms of its required input parameters, given output parameters, and other binding parametric details for runtime execution. The description also contains a short NL narrative of the overall behavior. A query dataset for this corpus was developed by three research assistants. The task for each of these three assistants was to formulate a *wh*-query for every service such that the query desire matches the given output of the service and query input matches the required input of the service. Since this task was done independently we observed that almost in all cases the syntactic structuring of the query for a given service by each assistant was different. The queries were simple, complex, and compound with an average of 90% query of the form complex and compound. Ideally, the extracted query desire by QCT should be semantically equivalent the output parameter of the corresponding web service specification. Based on this notion we have calculated CC-precision, CC-recall and CC-F1 measure for each of the three query datasets. From table 6 we observed that the average recall was 98.77%, average precision 100% and average F1 was 98.92%. The results clearly validate the earlier results with Microsoft Encarta 98.

## 6. CONCLUSION

In this paper we have a Description Logic based NL query formalization methodology. The motivation is to improve accuracy of answer extraction from NL documents using formal logic based reasoning. We have proposed the basic DL translation rules along with some of the important derived rules that cover different kinds of linguistic nuances. We found promising results while evaluating $DLQ_{S-Wh}M$ with MS Encarta query test set and a query dataset built on top of OWL-S TC v.4.0 dataset.

---

[6] http://projects.semwebcentral.org/projects/owls-tc/

**Table 5.1**

| Query Types | $N_{wh}$ | $N_{I-wh}$ | $N_{CI-wh}$ | CC-Re. (%) | CC-Pr. (%) | CC-F1 (%) |
|---|---|---|---|---|---|---|
| Simple Wh-Query | 676 | 642 | 642 | 94.97 | 100 | 97.42 |
| Complex Wh-Query | 147 | 140 | 140 | 95.23 | 100 | 97.55 |
| Compound Wh-Query | 69 | 64 | 64 | 92.75 | 100 | 96.23 |
| **Total** | 892 | 843 | 843 | 94.50 | 100 | 97.17 |

**Table 5.2**

| Query Types | $N_{wh}$ | $N_{I-wh}$ | $N_{CI-wh}$ | CC-Re. (%) | CC-Pr. (%) | CC-F1 (%) |
|---|---|---|---|---|---|---|
| How | 165 | 158 | 158 | 96.68 | 100 | 98.31 |
| What | 406 | 392 | 392 | 95.75 | 100 | 97.83 |
| When | 39 | 35 | 35 | 96.55 | 100 | 98.24 |
| Where | 85 | 82 | 82 | 89.74 | 100 | 94.59 |
| Which | 5 | 5 | 5 | 96.47 | 100 | 98.20 |
| Who | 143 | 143 | 143 | 100 | 100 | 100 |
| **Total** | 843 | 815 | 815 | 96.68 | 100 | 98.31 |

**Table 6**

| Query Types | $N_{wh}$ | $N_{I-wh}$ | $N_{CI-wh}$ | CC-Re. (%) | CC-Pr. (%) | CC-F1 (%) |
|---|---|---|---|---|---|---|
| NL Query$_{user1}$ | 1083 | 1000 | 1000 | 97.65 | 100 | 98.81 |
| NL Query$_{user2}$ | 1083 | 997 | 997 | 97.36 | 100 | 98.66 |
| NL Query$_{user3}$ | 1083 | 1010 | 1010 | 98.63 | 100 | 99.31 |
| **Total** | 3249 | 3007 | 3007 | 97.88 | 100 | 98.92 |